\documentclass{article}
\usepackage{nips14submit_e,times}
\nipsfinalcopy

\usepackage{times}
\usepackage{graphicx} 
\usepackage{subfigure} 

\usepackage{natbib}

\usepackage{algorithm}
\usepackage{algorithmic}

\usepackage{hyperref}

\usepackage{amsbsy}

\begin{document} 

\title{Lateral Connections in Denoising Autoencoders Support
Supervised Learning}

\author{Antti Rasmus \\ Aalto University, Finland \And
Harri Valpola \\ ZenRobotics Ltd., Finland \And
Tapani Raiko \\ Aalto University, Finland}


\maketitle

\begin{abstract} 
We show how a deep denoising autoencoder with lateral connections can be used
as an auxiliary unsupervised learning task to support supervised learning.
The proposed model is trained to minimize simultaneously the sum of supervised 
and unsupervised cost functions by 
back-propagation, avoiding the need for layer-wise pretraining. It improves
the state of the art significantly in the permutation-invariant MNIST 
classification task.

\end{abstract} 

\section{Introduction}

Combining an auxiliary task to help train a neural network was proposed by
\citet{suddarth1990rule}. By sharing the hidden representations 
among more than one task, the network generalizes better.
\citet{hinton2006reducing} proposed that this auxiliary task could be 
unsupervised modelling of the inputs.
\citet{Szummer2008semi-supervisedlearning} used autoencoder reconstruction
as auxiliary task for classification but performed the training layer-wise.

\citet{sietsma1991creating} proposed to corrupt network inputs with noise as 
a regularization method. Denoising autoencoders
\citep{vincent2010stacked} use the same principle to create unsupervised 
models for data.
\citet{Rasmus15arxiv} showed that modulated lateral connections in denoising
autoencoder change its properties in a fundamental way making it more suitable
as an auxiliary task for supervised training:
\begin{itemize}
\item Lateral connections allow the detailed information to flow directly
to the decoder relieving the pressure of higher layers to represent all 
information and allowing them to concentrate on more abstract features.
In contrast to a deep denoising autoencoder, encoder can 
discard information on the way up similarly
to typical supervised learning tasks discard irrelevant information.

\item With lateral connections,
the optimal model shape is pyramid like, i.e. the dimensionality
of the top layers is lower than the bottom layers, which is also true for
typical supervised learning tasks, as opposed to traditional denoising
autoencoders which prefer layers that are equal in size.
\end{itemize}
This paper builds on top the previous work and shows that using denoising 
autoencoder with lateral connections as an auxiliary task for supervised 
learning improves network's generalization capability as hypothesized by
\citet{valpola2015ladder}. The proposed method achieves
state-of-the-art results in permutation invariant MNIST classification task.




\section{Proposed Model}
\renewcommand{\l}{^{(l)}}
\newcommand{\x}{\mathbf{x}}
\newcommand{\n}{\emph{n}}
\renewcommand{\b}{\mathbf{b}}
\newcommand{\y}{\mathbf{y}}
\renewcommand{\t}{\mathbf{t}}
\newcommand{\e}{\mathbf{e}}
\newcommand{\f}[1]{f^{(#1)}}
\newcommand{\g}[1]{g^{(#1)}}
\newcommand{\p}[1]{\mathbf{p}^{(#1)}}
\renewcommand{\u}[1]{\mathbf{u}^{(#1)}}
\newcommand{\h}[1]{\mathbf{h}^{(#1)}}
\renewcommand{\b}[1]{\mathbf{b}^{(#1)}}
\newcommand{\z}[1]{\mathbf{z}^{(#1)}}
\newcommand{\W}[1]{\mathbf{W}^{(#1)}}
\newcommand{\V}[1]{\mathbf{V}^{(#1)}}

\begin{figure}[tbp]
\begin{center}
\centerline{\includegraphics[width=0.7\columnwidth]{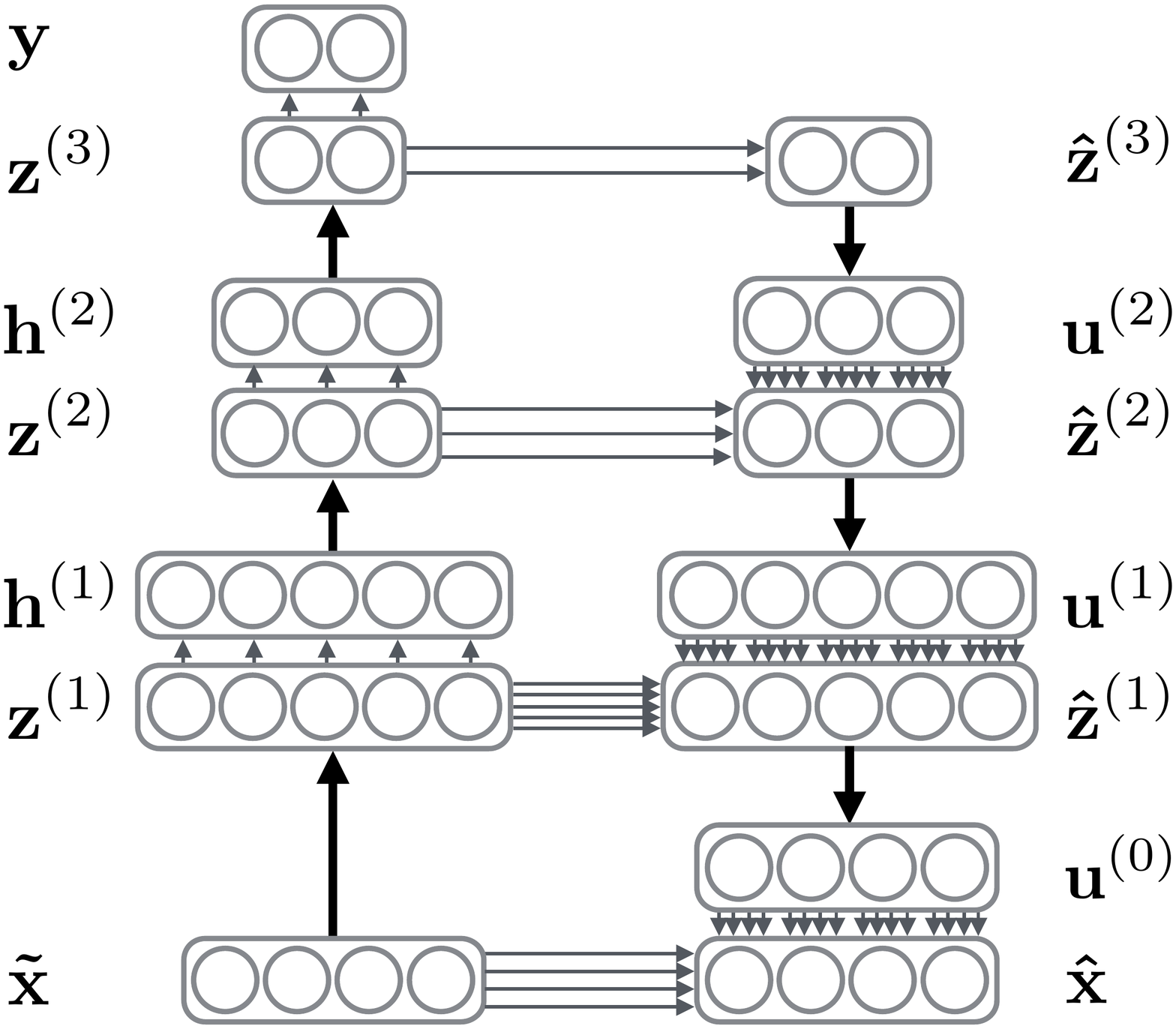}}
\caption{The conceptual illustration of the model when $L=3$.
Encoder path from $\tilde \x \to \y$ 
is a multilayer perceptron network, bold arrows indicating fully connected
weights $\W 1 \dots \W 3$ upwards and $\V 3 \dots \V 1$ downwards
and thin arrows neuron-wise connections.
$\z l$ are normalized preactivations, $\hat \z l$ their denoised versions,
and $\hat \x$ denoised reconstruction of the input. $\u l$ are projections
of $\hat \z {l+1}$ in the dimensions of $\z l$. $\h l$ are the activations
and $\y$ the class prediction.
}
\label{fig:model}
\end{center}
\end{figure}

The encoder of the autoencoder acts as the multilayer perceptron network for 
the supervised task so that the prediction is made in the highest layer of the
encoder as depicted in Figure~\ref{fig:model}.
For the decoder, we follow the model by \citet{Rasmus15arxiv} but with
more expressive decoder function and other minor modifications
described in Section~\ref{sec:model.unsuper}.

\subsection{Encoder and Classifier}

We follow \citet{Ioffe15arxiv} to apply batch normalization to each 
preactivation including the topmost layer in $L$-layer network
to ensure fast convergence due to reduced covariate shift. Formally, when 
input $\h 0 = \tilde \x$ and $l=1\dots L$
$$\z l = N_B(\W l \h {l-1})$$
$$h^{(l)}_i = \phi(\gamma^{(l)}_i(z^{(l)}_i + \beta^{(l)}_i))$$
where $N_B$ is a component-wise batch normalization $N_B(x_i) =
\frac{x_i - \hat\mu_{x_i}}{\hat\sigma_{x_i}}$,
where $\hat\mu_{x_i}$ and $\hat\sigma_{x_i}$ are estimates calculated from
the minibatch, $\gamma^{(l)}_i$ and $\beta^{(l)}_i$ are trainable  
parameters, and $\phi(\cdot)=\max (0,\cdot)$ is the rectification 
nonlinearity, which is replaced by the softmax for the output $\y=\h L$.

As batch normalization is reported to reduce the need of dropout-style
regularization, we only add isotropic Gaussian noise $n$ to the inputs,
$\tilde \x = \h 0 = \x + n$.

The supervised cost is average negative log probability of the targets 
$t(n)$ given the inputs $\mathbf{x}(n)$
$$C_\mathrm{class} = - \frac 1 N \sum_{n=1}^N \log P(Y=t(n)\mid 
\mathbf{x}(n)).$$

\subsection{Decoder for Unsupervised Auxiliary Task}
\label{sec:model.unsuper}

The unsupervised auxiliary task performs denoising similar to traditional 
denoising autoencoder, that is, it tries to match the reconstruction $\hat \x$
with the original $\x$.

Layer sizes in the decoder are symmetric to the encoder and corresponding
decoder layer $\hat \z l$ is calculated from lateral connection $\z l$
and vertical connection $\hat \z {l+1}$. 
Lateral connections are restricted so that each unit $i$ in an encoder layer
is connected to only one unit $i$ in the corresponding decoder layer, but
vertical connections are fully connected and projected to the same space
as $\z l$ by
$$\u l = \V {l+1} \hat \z {l+1},$$
and lateral neuron-wise connection for the $i$th neuron is
$$\hat z_i = a_{i1} z_i + a_{i2} \sigma(a_{i3} z_i + a_{i4})
+ a_{i5},$$
$$a_{ij} = c_{ij} u_i + d_{ij},$$
where superscripts $^{(l)}$ are dropped to avoid clutter, 
$\sigma(\cdot)$ is the sigmoid nonlinearity, and 
$c_{ij}^{(l)}$ and $d_{ij}^{(l)}$ are the trainable parameters.
This type of parametrization allows the network to use information from higher 
layer for any $a_{ij}^{(l)}$. 
The highest layer $L$ has $\u L = \mathbf 0$ and the lowest layer
$\hat \x = \hat \z 0$ and $\z 0 = \tilde \x$.

\citet[][Section 4.1]{valpola2015ladder} discusses how denoising functions 
represent corresponding distributions.
The proposed parametrization suits many different distributions,
e.g. super- and sub-Gaussian, and multimodal. Parameter $a_{i2}$ defines
the distance of peaks in multimodal distributions (also the ratio of
variances if the distribution is a mixture of two distributions with
the same mean but different variance).
Moreover, this kind of decoder function is able to emulate both the additive
and modulated connections that were analyzed by \citet{Rasmus15arxiv}. 

The cost function for unsupervised path is the mean squared error, $n_\x$
being the dimensionality of the data
$$C_\mathrm{reconst}= - \frac 1 N \sum_{n=1}^N \frac 1 {n_\x} ||\hat \x(n) - 
\x(n)||^2$$
The training criterion is a combination of the two such that multiplier $\eta$
determines how much the auxiliary cost is used, and the case $\eta=0$
corresponds to pure supervised learning:
$$C = C_\mathrm{class} + \eta C_\mathrm{reconst}$$

The parameters of the model include $\W l$, $\boldsymbol{\gamma}^{(l)}$, and 
$\boldsymbol{\beta}^{(l)}$ for the encoder, and $\V l$, $\mathbf{c}_{j}^{(l)}$, 
and $\mathbf{d}_{j}^{(l)}$ for the decoder.
The encoder and decoder have roughly the same number of parameters
because the matrices $\V l$ equal to $\W l$ in size.
The only difference comes from per-neuron parameters, which encoder
has only two ($\gamma_i$ and $\beta_i$), but the decoder has ten ($c_{ij}$ 
and $d_{ij}$, $j=1\dots 5$).

\section{Experiments}

In order to evaluate the impact of unsupervised auxiliary cost to the generalization
performance, we tested the model with MNIST classification task. We randomly split
the data into 50.000 examples for training and 10.000 examples for validation.
The validation set was used for evaluating the model structure and hyperparameters
and finally to train model for test error evaluation.
To improve statistical reliability, we considered the average of
10 runs with different random seeds. Both the supervised and unsupervised cost
functions use the same training data.

Model training took 100 epochs with minibatch size of 100, equalling to
50.000 weight updates. We used Adam optimization 
algorithm \citep{kingma2015adam} for weight updates adjusting the
learning rate according to a schedule where the learning rate is linearly reduced
to zero during the last 50 epochs starting from 0.002.
We tested two models with layer sizes 784-1000-500-10 and 784-1000-500-250-250-250-10,
of which the latter worked better and is reported in this paper. The best input noise 
level was $\sigma=0.3$ and chosen from $\{0.1, 0.3, 0.5\}$.
There are
plenty of hyperparameters and various model structures left to tune but we were
satisfied with the reported results.

\begin{figure}[tb]
\begin{center}
\centerline{\includegraphics[width=0.5\columnwidth]{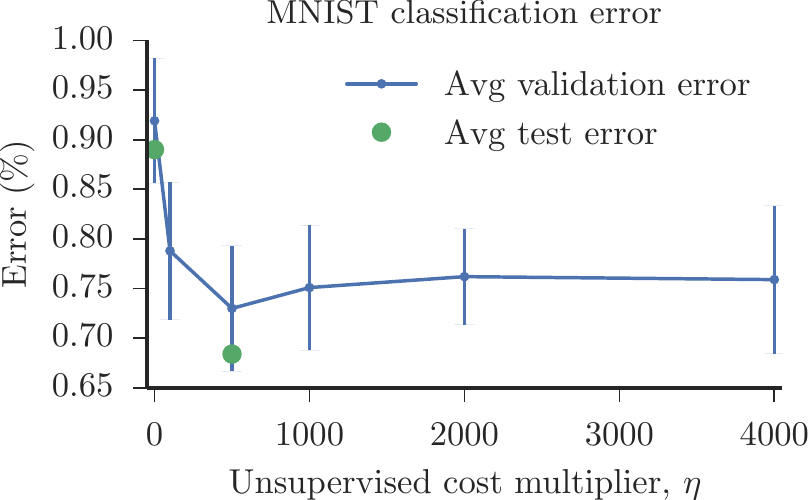}}
\caption{Average validation error as a function of unsupervised auxiliary cost multiplier $\eta$ and
average test error for the cases $\eta=0$ and $\eta=500$ over 10 runs.
$\eta=0$ corresponds to pure supervised training. Error bars show the sample 
standard deviation. Training included 50.000 samples for validation but
for test error all 60.000 labeled samples were used.}
\label{fig:denois}
\end{center}
\end{figure}

\subsection{Results}
Figure~\ref{fig:denois} illustrates how auxiliary cost impacts validation error by
showing the error as a function of the multiplier $\eta$. The auxiliary task is clearly
beneficial and in this case the best tested value for $\eta$ is 500.

The best hyperparameters were chosen based on the validation error results and then
retrained 10 times with all 60.000 samples and measured
against the test data. The worst test error was $0.72~\%$, i.e. 72 misclassified
examples, and the average $0.684~\%$ which is significantly lower than the previously
reported $0.782~\%$. For comparison, we computed the average test error for the
$\eta=0$ case, i.e. supervised learning with batch normalization, and got $0.89~\%$.

\section{Related Work}

Multi-prediction deep Boltzmann machine (MP-DBM) \citep{goodfellow2013multi} is 
a way to train a DBM with back-propagation through variational inference. The 
targets of the inference include both supervised targets (classification) and 
unsupervised targets (reconstruction of missing inputs) that are used in 
training simultaneously. The connections through the inference network are 
somewhat analogous to our lateral connections. Specifically, there are 
inference paths from observed inputs to reconstructed inputs that do not go all 
the way up to the highest layers. Compared to our approach, MP-DBM requires an 
iterative inference with some initialization for the hidden activations, 
whereas in our case, the inference is a simple single-pass feedforward 
procedure.

\begin{table}[t]
\begin{center}
 \begin{tabular}{ll}
   Method & Test error \\
   \hline
   SVM & 1.40 \% \\
   MP-DBM \cite{goodfellow2013multi} & 0.91 \% \\
   This work, $\eta=0$ & {0.89 \%} \\
   Manifold Tangent Classifier \cite{rifai2011manifold} & 0.81 \% \\
   DBM pre-train + Do \cite{srivastava2014dropout} & 0.79 \% \\
   Maxout + Do + adv \cite{goodfellow2015adver} & 0.78 \% \\
   This work, $\eta=500$ & \textbf{0.68 \%}
 \end{tabular}
 \end{center}
 \caption{A collection of previously reported MNIST test errors in
 permutation-invariant setting.
 Do: Dropout,
 adv: Adversarial training,
 DBM: deep Boltzmann machine.
 }
 \label{tab:results}
\end{table}

\section{Discussion}

We showed that a denoising autoencoder with lateral connections is
compatible with supervised learning using the 
unsupervised denoising task as an auxiliary
training objective, and achieved good results in MNIST classification
task with a significant margin to the previous state of the art.
We conjecture that the good results are due to supervised and unsupervised 
learning happening concurrently which means that unsupervised learning can 
focus on the features which supervised learning finds 
relevant.

The proposed model is simple and easy to implement with many existing feedforward
architectures, as the training is based on back-propagation from a simple cost 
function. It is quick to train and the convergence is fast, especially with
batch normalization. The proposed architecture implements complex 
functions such as modulated connections without a significant increase in the 
number of parameters.

This work can be further improved and extended in many ways. We are currently 
studying the impact of adding noise also to $\z l$ and including 
auxiliary layer-wise reconstruction costs $||\hat \z l - \z l||^2$, and working 
on extending these
preliminary experiments to larger datasets, to semi-supervised learning problems,
and convolutional networks.

\bibliography{example_paper}

\begin{thebibliography}{}

\bibitem[Goodfellow {\em et~al.}(2013)Goodfellow, Mirza, Courville, and
  Bengio]{goodfellow2013multi}
Goodfellow, I., Mirza, M., Courville, A., and Bengio, Y. (2013).
\newblock Multi-prediction deep {B}oltzmann machines.
\newblock In {\em Advances in Neural Information Processing Systems\/}, pages
  548--556.

\bibitem[Goodfellow {\em et~al.}(2015)Goodfellow, Shlens, and
  Szegedy]{goodfellow2015adver}
Goodfellow, I., Shlens, J., and Szegedy, C. (2015).
\newblock Explaining and harnessing adversarial examples.
\newblock {\em arXiv:1412.6572\/}.

\bibitem[Hinton and Salakhutdinov(2006)Hinton and
  Salakhutdinov]{hinton2006reducing}
Hinton, G.~E. and Salakhutdinov, R.~R. (2006).
\newblock Reducing the dimensionality of data with neural networks.
\newblock {\em Science\/}, {\bf 313}(5786), 504--507.

\bibitem[Ioffe and Szegedy(2015)Ioffe and Szegedy]{Ioffe15arxiv}
Ioffe, S. and Szegedy, C. (2015).
\newblock Batch normalization: Accelerating deep network training by reducing
  internal covariate shift.
\newblock {\em arXiv:1502.03167\/}.

\bibitem[Kingma and Ba(2015)Kingma and Ba]{kingma2015adam}
Kingma, D. and Ba, J. (2015).
\newblock Adam: A method for stochastic optimization.
\newblock {\em arXiv:1412.6980\/}.

\bibitem[Ranzato and Szummer(2008)Ranzato and
  Szummer]{Szummer2008semi-supervisedlearning}
Ranzato, M.~A. and Szummer, M. (2008).
\newblock Semi-supervised learning of compact document representations with
  deep networks.
\newblock In {\em Proceedings of the 25th International Conference on Machine
  Learning\/}, ICML '08, pages 792--799. ACM.

\bibitem[Rasmus {\em et~al.}(2015)Rasmus, Raiko, and Valpola]{Rasmus15arxiv}
Rasmus, A., Raiko, T., and Valpola, H. (2015).
\newblock Denoising autoencoder with modulated lateral connections learns
  invariant representations of natural images.
\newblock {\em arXiv:1412.7210\/}.

\bibitem[Rifai {\em et~al.}(2011)Rifai, Dauphin, Vincent, Bengio, and
  Muller]{rifai2011manifold}
Rifai, S., Dauphin, Y.~N., Vincent, P., Bengio, Y., and Muller, X. (2011).
\newblock The manifold tangent classifier.
\newblock In {\em Advances in Neural Information Processing Systems\/}, pages
  2294--2302.

\bibitem[Sietsma and Dow(1991)Sietsma and Dow]{sietsma1991creating}
Sietsma, J. and Dow, R.~J. (1991).
\newblock Creating artificial neural networks that generalize.
\newblock {\em Neural networks\/}, {\bf 4}(1), 67--79.

\bibitem[Srivastava {\em et~al.}(2014)Srivastava, Hinton, Krizhevsky,
  Sutskever, and Salakhutdinov]{srivastava2014dropout}
Srivastava, N., Hinton, G., Krizhevsky, A., Sutskever, I., and Salakhutdinov,
  R. (2014).
\newblock Dropout: A simple way to prevent neural networks from overfitting.
\newblock {\em The Journal of Machine Learning Research\/}, {\bf 15}(1),
  1929--1958.

\bibitem[Suddarth and Kergosien(1990)Suddarth and Kergosien]{suddarth1990rule}
Suddarth, S.~C. and Kergosien, Y. (1990).
\newblock Rule-injection hints as a means of improving network performance and
  learning time.
\newblock In {\em Neural Networks\/}, pages 120--129. Springer.

\bibitem[Valpola(2015)Valpola]{valpola2015ladder}
Valpola, H. (2015).
\newblock From neural {P}{C}{A} to deep unsupervised learning.
\newblock In {\em Advances in Independent Component Analysis and Learning
  Machines\/}. Elsevier.
\newblock Preprint available as arXiv:1411.7783.

\bibitem[Vincent {\em et~al.}(2010)Vincent, Larochelle, Lajoie, Bengio, and
  Manzagol]{vincent2010stacked}
Vincent, P., Larochelle, H., Lajoie, I., Bengio, Y., and Manzagol, P.-A.
  (2010).
\newblock Stacked denoising autoencoders: Learning useful representations in a
  deep network with a local denoising criterion.
\newblock {\em The Journal of Machine Learning Research\/}, {\bf 11},
  3371--3408.

\end{thebibliography}
\bibliographystyle{natbib}

\end{document}